\documentclass[11pt,a4paper]{article}
\usepackage[hyperref]{eacl2021}
\usepackage{times}
\usepackage{latexsym}

\usepackage{graphicx}
\usepackage{amsmath}
\usepackage{makecell}
\usepackage[T1]{fontenc}

\usepackage[inline]{enumitem}
\setlist[itemize]{leftmargin=3mm}

\newcommand{\refoptauto}{\ref{tb:option_auto}}
\newcommand{\refoptgenmethod}{\ref{sec:baselines}}

\newcommand{\refAppendixImplement}{Appendix \ref{sec:implement_details}}
\newcommand{\refAppendixExp}{Appendix \ref{sec:experiment_details}}
\newcommand{\refAppendixMoreExp}{Appendix \ref{sec:more_experiment}}
\newcommand{\refAppendixMoreExample}{Appendix \ref{sec:more_example}}

\usepackage[ruled,vlined]{algorithm2e}

\usepackage{microtype}

\usepackage{multirow}

\usepackage{amsmath,amsfonts,bm}

\def\eqref#1{equation~\ref{#1}}

\def\1{\bm{1}}

\DeclareMathAlphabet{\mathsfit}{\encodingdefault}{\sfdefault}{m}{sl}
\SetMathAlphabet{\mathsfit}{bold}{\encodingdefault}{\sfdefault}{bx}{n}

\DeclareMathOperator*{\argmin}{arg\,min}

\aclfinalcopy 
\def\aclpaperid{995} 

\title{Changing the Mind of Transformers \\ for Topically-Controllable Language Generation}

\author{Haw-Shiuan Chang \ \ \ Jiaming Yuan \ \ \ Mohit Iyyer \ \ \ Andrew McCallum \\
CICS, University of Massachusetts Amherst\\
  {\tt hschang@cs.umass.edu,jiamingyuan@umass.edu, } \\ {\tt \{mccallum,miyyer\}@cs.umass.edu}
}

\date{}

\begin{document}
\maketitle
\begin{abstract}
Large Transformer-based language models can aid human authors by suggesting plausible continuations of text written so far. However, current interactive writing assistants do not allow authors to guide  text generation in desired topical directions. To address this limitation, we design a framework that displays multiple candidate upcoming topics, of which a user can select a subset to guide the generation. Our framework consists of two components: (1) a method that produces a set of candidate topics by predicting the centers of word clusters in the possible continuations, and (2) a text generation model whose output adheres to the chosen topics. The training of both components is self-supervised, using only unlabeled text. Our experiments demonstrate that our topic options are better than those of standard clustering approaches, and our framework often generates fluent sentences related to the chosen topics, as judged by automated metrics and crowdsourced workers.

\end{abstract}

\section{Introduction}

Recently, Transformer-based language models (LMs) have achieved impressive performance in language generation tasks~\citep{radford2019language,dai2019transformer} such as open-domain story generation~\citep{ see2019massively}.
When writing with the LM,  users often desire an intuitive and effective way to control what a LM is going to generate~\cite{keskar2019ctrl}. To address this need, interactive writing assistants provide options to reveal possible developments of the story and generate continuations guided by the user-selected options. 

Interactive writing assistants have wide applications in creative writing~\citep{roemmele2015creative,clark2018creative,akoury2020storium}, education~\citep{luo2015review}, and gaming~\citep{AI_Dungeon}. Nevertheless, the existing systems' options usually do not provide fine-grained control and/or require substantial human labor.
In some prior work~\citep{keskar2019ctrl, tu2019generating}, users choose among a static set of predefined attributes (e.g., sentiment) that only provide coarse-grained control. Other work~\citep{roemmele2015creative,clark2018creative} presents users with multiple generated continuations, which requires substantial reading effort and might not contain topics that users want to see. Finally, options could be nodes in a plot graph that are handcrafted~\citep{luo2015review} or derived from a collaboration between humans and machine~\citep{li2013story}, but such choices are usually limited due to the high cost of preparing the options.


\begin{figure}[t!]
\centering
\includegraphics[width=1\linewidth]{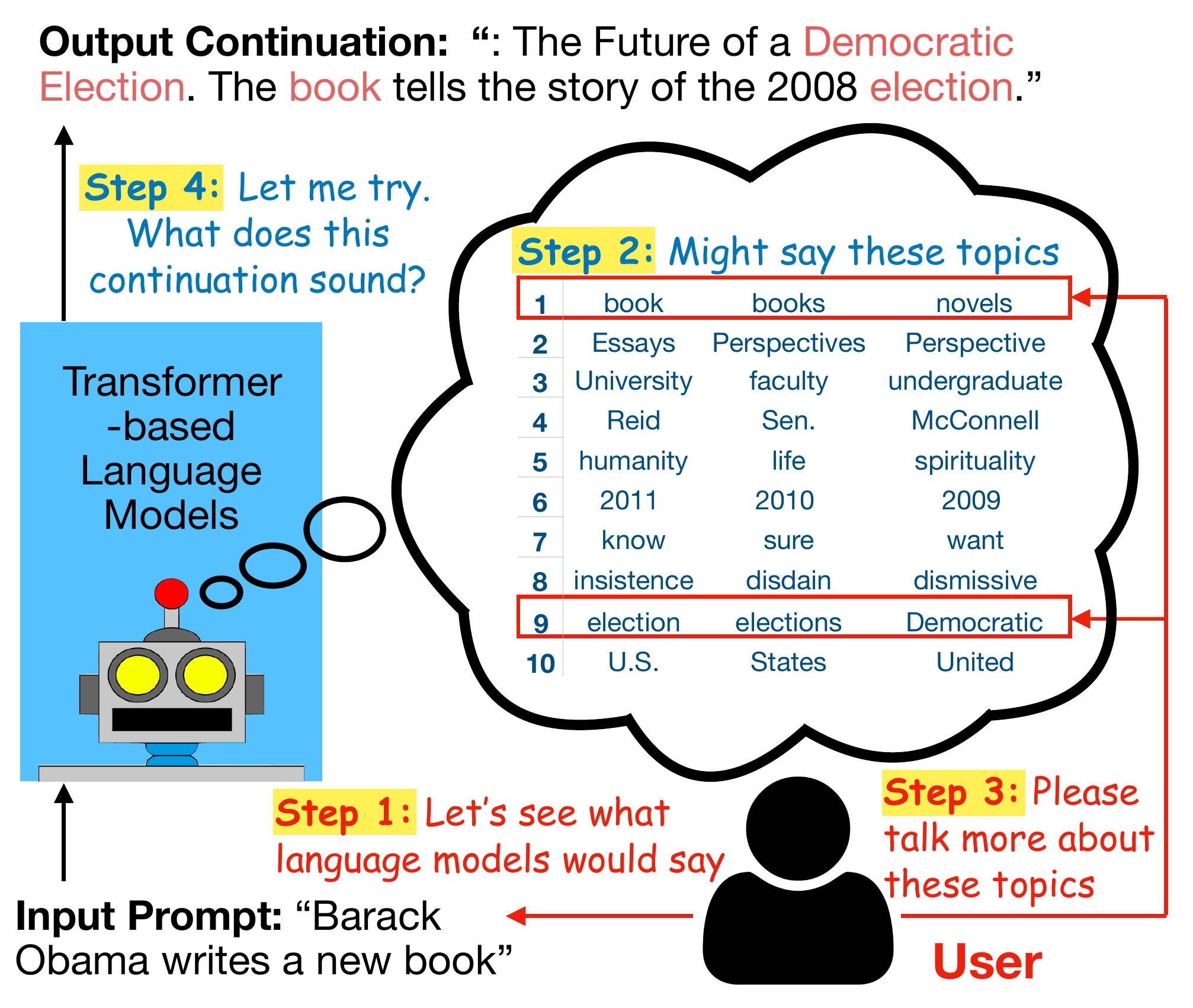}
\caption{Given an input prompt, the Transformer-based LM provides $K=10$ topics that might be mentioned next and each topic is represented by $M=3$ words. The user could guide the generation process by choosing a subset of topics.}
\label{fig:first_page}
\end{figure}

\begin{figure*}[t!]
\centering
\includegraphics[width=1\linewidth]{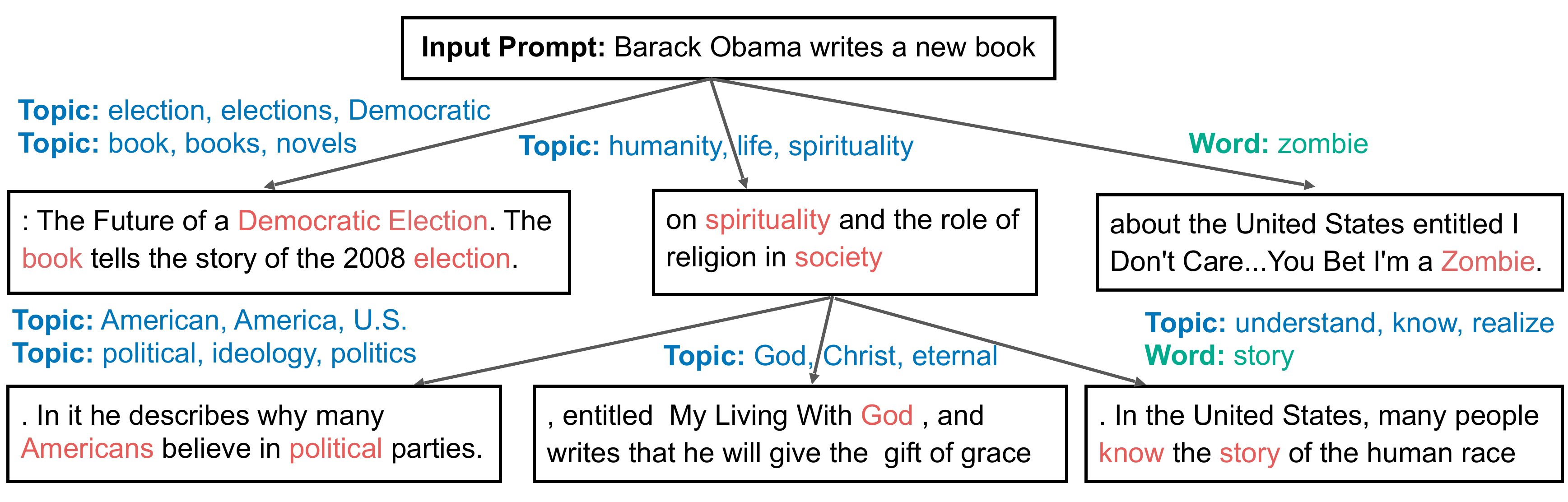}
\caption{Examples of our generated options and continuations. We highlight the words in the continuation that are related to the chosen topics or to the specified word.
}
\label{fig:Examples}
\end{figure*}

To address these limitations, we propose an interactive writing framework that provides a set of topics and guides the text generation by the user-chosen topics. The topic options are generated dynamically based on the input prompt to provide fine-grained control, and our models are self-supervised without the need to define the attributes or collect annotations.
As depicted in Figure~\ref{fig:first_page}, a user can peek at the most probable $K$ topics (shown as bags of words) appearing after the input prompt and control the generation by choosing the topics. 

In Figure~\ref{fig:Examples}, we compare multiple generated sentences conditioned on different chosen topic(s) or specified word(s). For example, if the user chooses a topic about \textit{humanity}, \textit{life}, and \textit{spirituality}, our system continues the input prompt ``\textit{Barack Obama writes a new book}'' with ``\textit{on spirituality and the roles of religion in society}''. Then, we can use the generated text as the new input prompt and update the set of topics to include other more relevant topics such as \textit{God}, \textit{Christ}, and \textit{eternal}. The process can be repeated to create a plot tree. 



A user can also control the generation by specifying word(s) if the user wants to see the words that are not in the topic list or seeks a transition to a word that is not directly related to the input prompt. 
For example, a user can ask our system to generate a sentence about \textit{zombie}. Consequently, the continuation of \textit{``Barack Obama writes a new book''} becomes \textit{``about the United States entitled I Don't Care...You Bet I'm a Zombie''}.



The system is realized by two components: an option generator and a conditional text generator. Given a prompt, the option generator suggests a set of $K$ topics. After a user chooses a subset of the topics and specifies some words, the embedding of every word or topic will guide the conditional text generator to produce the continuation that is both consistent with the existing prompt and relevant to the chosen topics and words.  



Both components are self-supervised and use pretrained GPT2 models~\citep{radford2019language} to encode the input prompt. 
During training, the option generator predicts the cluster centers of future words, which are in the continuation of the prompt, based on the contextualized embeddings from GPT2.
The conditional text generator fine-tunes GPT2 to predict the next words given the prompt and a few subsequent words. Since both components' input and output only come from the prompt and its continuation, training the system only requires a raw corpus, word tokenizers, and a list of stop words. 
This makes the proposed method suitable for open-domain story generation and easily being fine-tuned for a specific domain.



In experiments, we demonstrate that our system recommends high-quality topics and often generate sentences that follow the chosen topics. We compare our option generator with global topic models such as LDA~\citep{blei2003latent} or local topic models such as clustering the words in the input prompt. The results show that the proposed method generates significantly more topics that are plausible and promote the narrative. Moreover, we compare our conditional text generator with PPLM (Plug and Play Language Models)~\citep{dathathri2019plug} and demonstrate that our generation is more fluent and relevant to the chosen topics. Our code is available at \url{https://github.com/iesl/interactive_LM}.















\section{Method}

\begin{figure}[t!]
\centering
\includegraphics[width=1\linewidth]{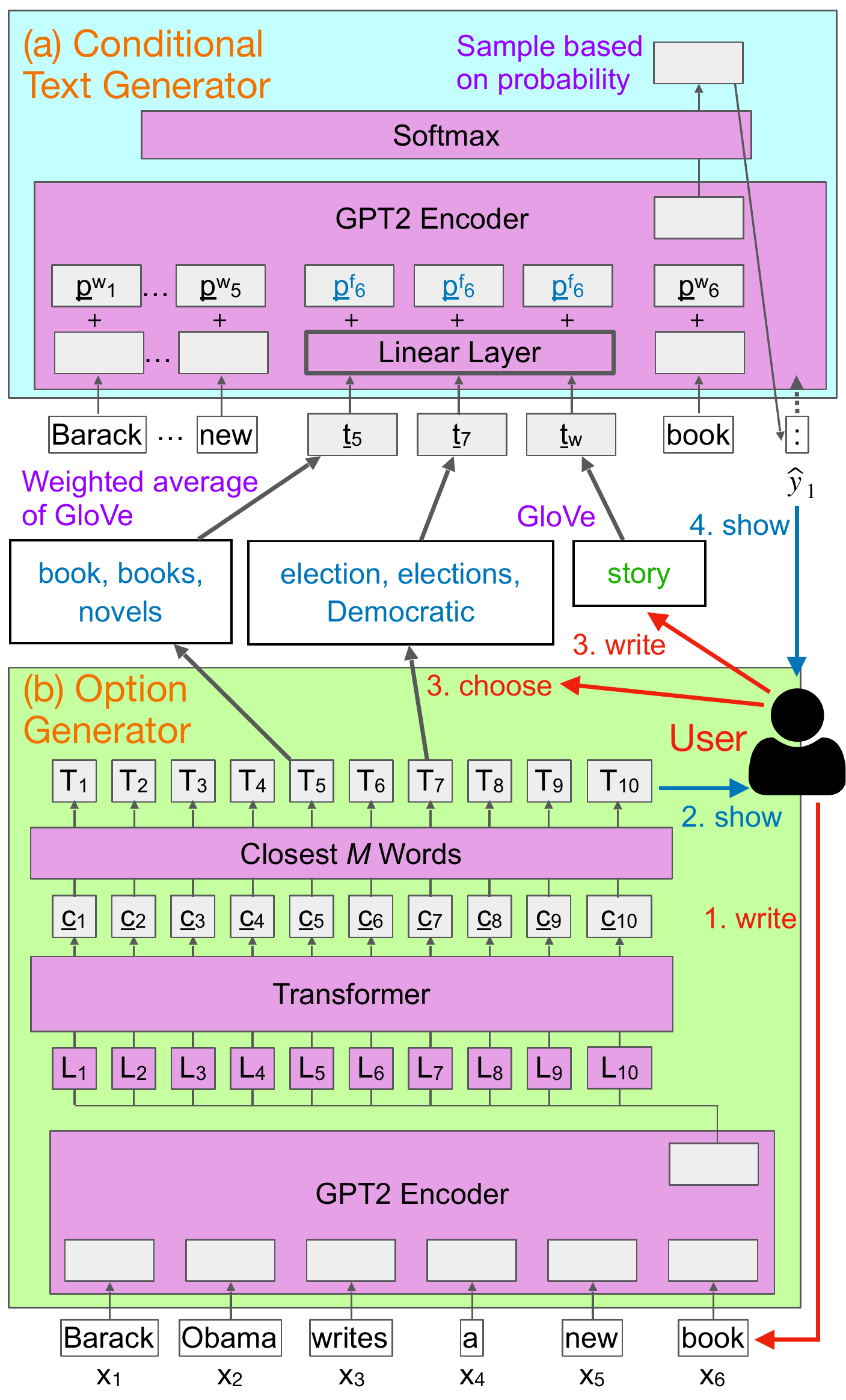}
\caption{Our model architectures for (a) conditional text generator and (b) option generator. During testing, the information flows from the bottom to the top.}
\label{fig:Testing}
\end{figure}

The proposed framework consists of two components: option generator and conditional text generator. In Figure~\ref{fig:Testing}, we illustrate the two components and their interaction. First, given the prompt $x_1,...,x_I$ inputted by a user, the option generator at the bottom of the figure outputs $K$ topics. After the user chooses two topics about \textit{book} and \textit{election} and specifies one extra word \textit{story}, the topics and word are passed to our text generator as the generation guidance. Accordingly, the generator continues to write the next token $\widehat{y}_1$.\footnote{The framework is flexible. For example, the GPT2 encoders in the two components could be shared.
Besides topics, the option generator could be extended to predict likely attributes in the continuation such as positive sentiment and event frames~\cite{tu2019generating} if the corresponding label data are available in the training corpus.}

In the following subsections, we introduce our model designs and the way to train each component. More implementation details are described in \refAppendixImplement.








\begin{figure*}[t!]
\centering
\includegraphics[width=1\linewidth]{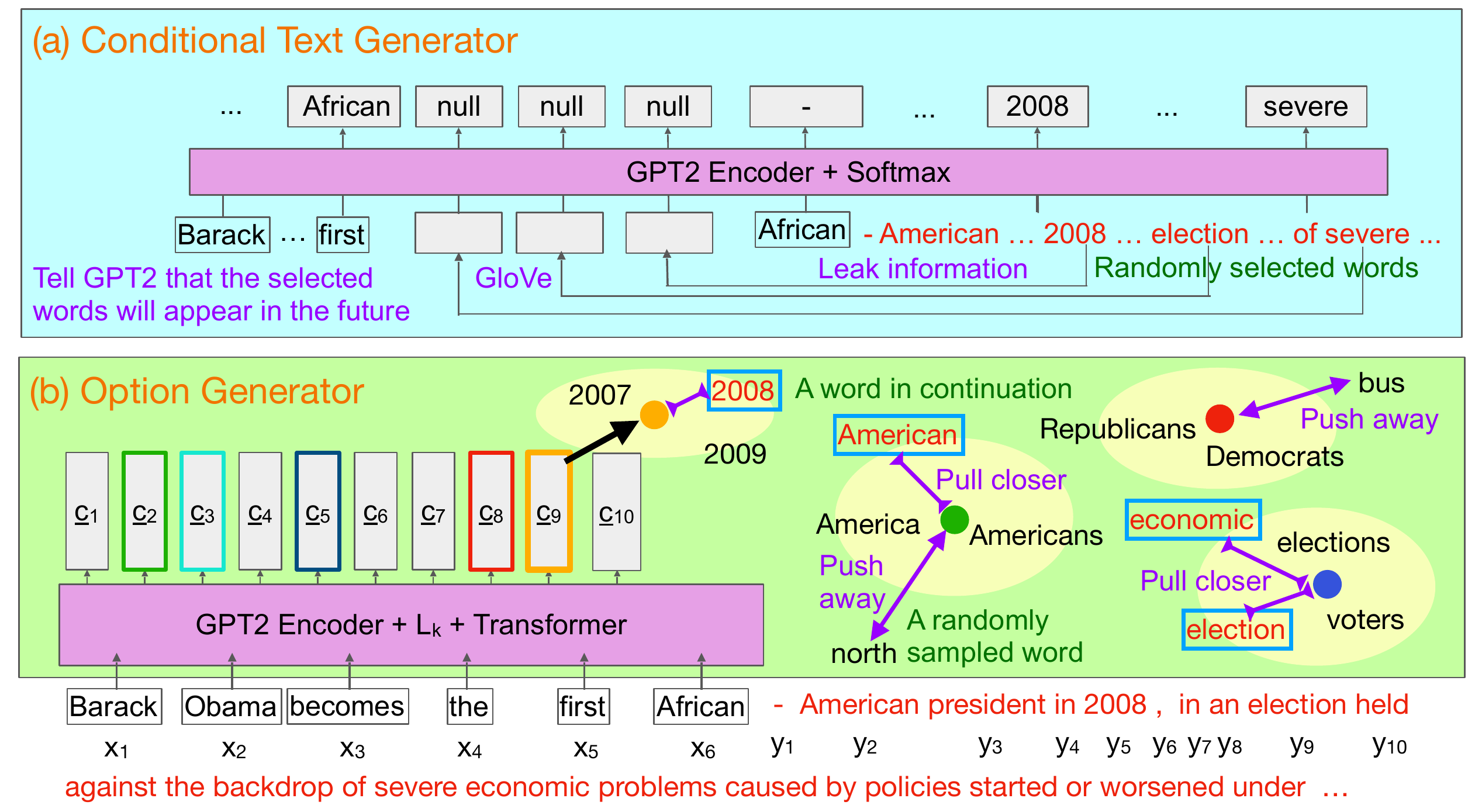}
\caption{Training our two components using the same sentence. (a) We randomly pick $n=3$ words in the actual continuation as our conditions for the text generator, and the null labels mean their predicted probabilities are ignored in our loss. (b) We visualize $5$ out of $K=10$ generated topics in a normalized GloVe space. Red words are the ones that appear in the continuation and pull the nearby cluster centers closer during training.}
\label{fig:Training}
\end{figure*}

\subsection{Option Generator}
When we do not have labeled attributes in a corpus, we can create options by clustering all the words in a corpus into topics~\citep{tu2019generating}. The clustering could be done by topic modeling approaches such as LDA~\citep{blei2003latent}. The resulting topics are static (i.e., the clustering is performed globally without considering the prompt).
However, the prompt might have a narrow focus and 
the related words of interest are all clustered into a single topic. 

A simple remedy is to cluster only the words in the prompt rather than all the words in the corpus. The topics are created dynamically and locally given a prompt and can capture more fine-grained aspects in the continuations. However, the topics derived from the prompt might provide less inspiration
because the users have seen the prompt. Another major drawback of the approach is that the generated topics might encourage the LM to generate repetitive sentences or make a narrative circle inside a loop. 

Motivated by the challenges, we propose an option generator that predicts the cluster centers based on the prompt instead of clustering the words in the prompt during testing. 


\subsubsection{Model Prediction}
The goal of our option generator is to predict the $K$ cluster centers of words in the possible continuations and use the cluster centers as the topics user could choose from. As in Figure~\ref{fig:Testing} (b), the option generator uses GPT2 to encode the input prompt $x_1,...,x_I$ and passes the output embedding to $K$ different linear layers $L_1,...,L_K$. To model the dependency of clusters, a Transformer~\citep{vaswani2017attention} takes the $K$ embeddings as input and predicts the cluster centers $\underline{c}_1,...\underline{c}_K$ in GloVe~\citep{glove} space. During testing, each predicted cluster center is normalized by its L2 norm, and we use the $M$ closest words in the normalized GloVe space to represent the topic $T_i$, which users can choose. 

We choose to learn the cluster centers in GloVe space rather than GPT2 or BERT~\citep{BERT} space because the non-contextualized word embeddings
are easier to visualize. Users can easily understand the meaning of a cluster center by seeing nearby words. We normalize GloVe space in this work to make the squared L2 distance equal to twice the cosine distance between two embeddings. 

Our architecture is similar to the one in~\citet{NSD}, but we use a pretrained GPT2 encoder rather than train a BERT-like Transformer from scratch. Another difference is that we ignore the connection between the second Transformer and the output of GPT2 to save GPU memory for handling a longer input prompt.






\subsubsection{Model Training}
In Figure~\ref{fig:Training} (b), we visualize our training procedure. For each input prompt in the training corpus, we run a forward pass through the Transformers and get predicted cluster centers $\underline{c}_1,...\underline{c}_K$. Next, we collect 50 words in the continuation (except stop words) as positive examples and match the words with cluster centers as in the E-step of the EM algorithm~\citep{dempster1977maximum}. We minimize the distances between the centers and their nearby positive examples by backpropagating the gradients through the matching and updating our Transformer models. Furthermore, we randomly sample some words as negative examples and maximize the distances between the cluster centers and their nearby embeddings from negative examples.




Using Figure~\ref{fig:Training} (b) as an example, the orange cluster center is pulled closer toward the embedding of \textit{2008}, which appears in the continuation. The green cluster center is pushed away from the embedding of \textit{north}, a randomly sampled word. Since each output embedding $\underline{c}_k$ is pulled by only the nearby embeddings of words in the continuation, the output embedding will naturally become the cluster center of the nearby continuation word embeddings.
Notice that the related topics like \textit{Democrats} and \textit{Republicans} are not observed in the prompt and continuation, but our model can predict a red cluster center close to them because the model can learn from other similar input prompts whose continuation mentions words like \textit{Democrats}. 

\citet{NSD} discover that non-negative sparse coding (NNSC)~\citep{hoyer2002non} could encourage the Transformers to predict more diverse and relevant topics compared with Kmeans, so we adopt NNSC as our clustering loss, and its formulation could be found in~\citet{NSD}.


\subsection{Conditional Text Generator}
After the user chooses topic(s) or specifies word(s), each topic or word is converted to a GloVe embedding. The component aims to generate the text given the input prompt and the GloVe embeddings of the topics or words we prefer to see in the continuation.

Users only see the $M$ words closest to the $k$th predicted cluster center $\underline{c}_k$ from our option generator, 
so we compute the $k$th topic embedding as 
\begin{align}
    \underline{t}_k = \frac{ \sum_{m=1}^M \text{cos}(\underline{e}^{w}_m,\underline{c}_k )\underline{e}^{w}_m }{||\sum_{m=1}^M \text{cos}(\underline{e}^{w}_m,\underline{c}_k )\underline{e}^{w}_m||}, 
    \label{eq:topic_embedding}
\end{align}
where $\underline{e}^{w}_m$ is the normalized GloVe embedding of the $m$th closet word and $\text{cos}(\underline{e}^{w}_m,\underline{c}_k )$ is the cosine similarities between the $m$th word embedding and the embedding $\underline{c}_k$.





\subsubsection{Model Prediction}
During testing, the topic embeddings $\underline{t}_k$ or embedding of the specified words are inserted into GPT2 encoder before $x_I$, the last word piece in the prompt. The inserted embeddings nudge the GPT2 to generate the sentences containing the desired words with a higher probability.  

As Figure~\ref{fig:Testing} (a) shows, the GloVe embeddings are first passed through a linear layer to make their dimension become the same as the hidden state size of GPT2. Then, the transformed embeddings are added with special positional embeddings $\underline{p}^f_I$, which are different from those for the prompt $\underline{p}^w_i$. The special positional embedding tells GPT2 that the inserted embeddings have a different meaning and where the conditional generation starts.

The GPT2 encoder's output goes through a softmax layer, which computes the probability of each token being observed as the first word piece in the continuation $y_1$. We adopt top-k sampling~\citep{fan2018hierarchical}, which reduces the chance of sampling words with low probability, to pick the next word, and autoregressively sample one token $\widehat{y}_o$ at a time to generate the continuation $\widehat{y}_1,...,\widehat{y}_O$.






\subsubsection{Model Training}
We train the generator using the continuation of a prompt and some randomly selected non-stop words in the continuation as its generation conditions. Since the continuation contains the randomly-selected words, the generator would be heavily penalized if it ignores the conditions by assigning low probabilities to the selected words in all the continuation positions.

An example is illustrated in Figure~\ref{fig:Training} (a). Given an input prompt in the training set, we randomly pick a number $n$ from $0$ to $K$ and sample $n$ words from the next $O=25$ words (except stop words). Next, the normalized GloVe embeddings of $n$ words are inserted to the GPT2 encoder before the last word piece in the prompt, and we ignore the output probabilities corresponding to the inserted positions during training. To speed up the training, we conduct the future word insertion in multiple positions of each training text sequence. 

We insert the future words just before the text that might contain the words rather than at the beginning as in the classic seq2seq model, because we do not want the model to learn to generate the continuation based on the future topics that have not yet be specified by the users (e.g., The GPT2 should not know that it will see \textit{election} in the future when it learns to generate \textit{Barack Obama ...} during training).

By allowing the LM to see the upcoming words earlier, we leak partial label information to the LM input. Consequently, GPT2 learns to utilize the information and generate the sentence containing the desired words to achieve a lower perplexity loss. Notice that the training method allows us to specify our topical preference without significantly scarifying generation efficiency and fluency, but it cannot guarantee to generate all the desired topics, especially when we specify multiple ones.

One concern of the method is that the LM cannot see all possible sets of topics or words users might specify during training. Besides, each GloVe embedding used to supervise LM comes from a single word, but we ask the LM to condition on average GloVe embedding of the top $M$ words during testing. Nevertheless, we observe that the LM is often able to generalize well in our experiments because similar words have similar GloVe embeddings, lots of training instances could be easily prepared by the self-supervised method, and our option generator usually provides the topics mentioned in the continuation in our training corpus.










\section{Experiments}

We evaluate two components separately, and both evaluations include automated metrics and human judgment. Throughout the evaluation, the number of topics $K=10$ and the length of generations is 50 word pieces. We find that fixing $K=10$ works well in our experiments. If the possible continuations cover more than 10 topics, our option generator tends to output the important topics. If they cover fewer topics, our option generator tends to output the related topics that are not explicitly mentioned in the prompt or the duplicated topics. More experiment setup details could be found in \refAppendixExp. 

\subsection{Datasets}

We use 90\% of English Wikipedia 2016 as our training set for both components, 5\% as our validation set to determine the hyperparameters such as the number of epochs, and the remaining 5\% as our test set to perform the automated evaluation. 

For human evaluation, we collect labels from Amazon Mechanical Turk (MTurk). We randomly sample sentences from the training set of STS benchmark (STSb)~\cite{sts} as our input prompts. Compared with Wikipedia, the sentences from STSb are easier to understand for annotators because a large portion of sentences in Wikipedia involves terminologies, depends on a longer context, or might even just be a list of names.

In STSb, we sample 24 sentences as our prompts, and each method generates one continuation for each input prompt. Each generated continuation or topics will be scored by three different workers.







\subsection{Option Generator Evaluation}
We evaluate the topics from different option generators by judging whether the topics will appear in the continuation and whether the topics would promote the narrative. The goal is to have topics that are relevant and provide new information. The topics that are too similar to the prompt words might be redundant and not helpful because the users have already seen the prompt.






\subsubsection{Automatic Evaluation Metrics}
\label{sec:option_auto_metrics}
\begin{itemize}
\setlength\itemsep{-0.3em}
\item \textbf{Sim}: If the generated topics $T$ can help users to write the continuation, the embedding of every non-stop word in the actual continuation should be similar to the embeddings of a generated topic. Thus, we compute
\begin{align}
    \text{Sim}(\bar{Y},T) = \sum\limits_{o=1}^{O'} \max\limits_{k=1}^K (\underline{t}_k)^T \underline{e}^{\bar{y}}_o,
\end{align}
where $\bar{Y}=\{\bar{y}_o\}_{o=1}^{O'}$ is a set of non-stop words in the continuation and $O'=25$. $\underline{t}_k$ is the normalized embedding of $k$th topic in $T$ from \eqref{eq:topic_embedding} and $\underline{e}^{\bar{y}}_o$ is the $o$th word in $\bar{Y}$.  

\item \textbf{Sim Short}: When computing Sim, we use the input prompts containing around 180 words on average.
To examine the topic quality at the start of writing, where the authors might need assistance the most, we also report Sim$(\bar{Y},T)$ on short input prompts (with 35 words on average).

\item \textbf{Sim Diff}: The options that are helpful to users should be sufficiently different from the words in the input prompt to promote the narrative and avoid generating repeated content. Thereby, we also evaluate methods using Sim Diff = Sim$(\bar{Y},T)$ - Sim$(\bar{X},T)$, where $\bar{X}=\{\bar{x}_i\}_{i=1}^{I'}$ are the non-stop words in the input prompt.

\end{itemize}



\begin{table}[t!]
\scalebox{0.85}{
\begin{tabular}{|c|c|ccc|}
\hline
Scope & Method &  Sim & Sim Short & Sim Diff \\  \hline
\multirow{3}{*}{Global} & Sample & 14.63 & 14.42 &  \textbf{0.16}   \\
& LDA & 36.86 & 36.02 & -2.82   \\
& Kmeans & \textbf{40.65} & \textbf{39.91} & -3.40  \\ \hline
\multirow{4}{*}{Local} & Sample & 41.50 & 41.23 & -12.51 \\
& NNSC & 46.70 & 42.80 & -15.94  \\
& Kmeans & 47.94 & 43.89 & -16.12  \\
& Ours & \textbf{48.38} & \textbf{46.29} & \textbf{0.45}  \\ \hline 

\end{tabular}
}
\centering
\caption{Comparison of the option generators using automatic metrics. The best numbers within each scope are highlighted.}
\label{tb:option_auto}
\end{table}

\begin{table}[t!]
\scalebox{0.75}{
\begin{tabular}{|c|c|ccc|}
\hline
Scope & Method & L & TP & L\&TP \\ \hline
\multirow{2}{*}{Global} & LDA & 5.76 $\pm$ 0.50 & 6.24 $\pm$ 0.33 & 5.26 $\pm$ 0.31 \\ 
 & Kmeans & 6.94 $\pm$ 0.36 &  6.13 $\pm$ 0.30 &  5.96 $\pm$ 0.31 \\ \hline
\multirow{2}{*}{Local} & Kmeans & \textbf{8.65} $\pm$ 0.16 & 5.31 $\pm$ 0.50 & 5.14 $\pm$ 0.50 \\
& Ours & 7.85 $\pm$ 0.25 & \textbf{6.96} $\pm$ 0.26 & \textbf{6.75} $\pm$ 0.28 \\ \hline

\end{tabular}
}
\centering
\caption{ Comparison of option generators using human judgment (mean $\pm$ standard error). L and TP refer to likelihood and topic promotion, respectively.}
\label{tb:option_human}
\end{table}

\begin{table*}[t!]
\scalebox{0.6}{
\begin{tabular}{|cccc|cccc|cccc|}
\hline
\multicolumn{2}{|c|}{ {\Large \Gape[3pt][1pt]{Input Prompt}} } &  \multicolumn{10}{|c|}{ {\Large The study also found that skin cancer nearly tripled in Norway and Sweden since the 1950s.} }  \\ \hline
\multicolumn{4}{|c|}{ {\Large \Gape[3pt][1pt]{LDA-global} } } & \multicolumn{4}{|c|}{ {\Large Kmeans-local} } & \multicolumn{4}{|c|}{ {\Large Ours} } \\ \hline
1& population, households &6& company, companies &1& Norway, Sweden & 6& also, however &  1& research, scientific &6& 1980s, 1970s \\
2& patients, treatment &7& Norwegian, Norway &2& tripled, doubled & 7& since, Since &  2& tissues, tissue &7& even, though \\
3& psychology, research &8& story, book &3& nearly, almost & 8& Sweden, Finland &  3& patients, diagnosis &8& susceptibility, pathogenic \\
4& police, prison &9& hospital, Hospital& 4& cancer, skin & 9& study, studies  & 4& DNA, gene &9& decreased, increased \\
5& chemical, carbon &10& Icelandic, Iceland &5& 1950s, 1940s & 10& found, discovered &  5& orange, purple &10& Sweden, Norway \\ \hline


\end{tabular}
}
\centering
\caption{Comparison of all $K$ topics for the input prompt using $M=2$ words closest to each topic. }
\label{tb:generated_topics}
\end{table*}

\begin{table*}[t!]
\scalebox{0.75}{
\begin{tabular}{|cc|c|}
\hline
\multicolumn{2}{|c|}{Input Prompt} &  The study also found that skin cancer nearly tripled in Norway and Sweden since the 1950s. \\ \hline
\multicolumn{2}{|c|}{Generator} & \multirow{2}{*}{Generated Text}  \\ \cline{1-2}
Option & Text &   \\ \hline
LDA-global & Ours &   A study of the Norwegian police has confirmed the cancer case.  The law in Norway was the subject of the  \\
Kmeans-local & Ours &  The study also found that skin cancer nearly tripled in Norway and Sweden since the 1950s. As well, skin	 \\
Ours & PPLM & In this study, a study was conducted conducted in Italy and in Finland. From the 1990s to the 1970s, there \\
None & GPT2 & The study also revealed that only 20\% of the deaths in Norway were caused by a sudden cardiac response   \\
Ours & Ours &  Recent studies have shown that melanin causes a decrease in genetic susceptibility in people in Norway, \\ \hline

\end{tabular}
}
\centering
\caption{The continuations that are generated by conditioning on all of $K$ topics from different option generators. The input prompt comes from STSb.
}
\label{tb:generated_text_examples}
\end{table*}

\subsubsection{Human Evaluation}
Our questionnaire shows the prompt and asks which generated topics are likely to appear in a reasonable continuation and which topics promote the narrative. For each method, we report the average number of its topics that are likely to appear (L), promote the topic (TP), and both (L\&TP). For example, an MTurk worker is shown three topics generated by a method given a prompt: $ABC$. The worker thinks $A$ is likely to appear in the continuation and $AB$ promote the topic. Then, L=$|\{A\}|$=1, TP=$|\{AB\}|$=2, and L\&TP=$|\{A\}\cap\{AB\}|$=$|\{A\}|$=1 for this prompt.

\subsubsection{Option Generator Baselines}
\label{sec:baselines}
We compare our generator with two types of methods.\footnote{Another alternative is to generate many continuations and cluster the words in the generation. However, the method takes time, which might be prohibited by limited computational resources and the real-time interaction requirement.} The first type performs the clustering globally and selects the most relevant topics to the input prompt from the static set of clusters. We cluster all the words into $J=150$ topics by LDA~\citep{blei2003latent} (\textbf{LDA-global}) and into $J=1000$ topics by Kmeans on the normalized GloVe embedding space~\cite{tu2019generating} (\textbf{Kmeans-global}). We also randomly sample $K$ words from the whole vocabulary as our cluster centers (\textbf{Sample-global}). 


Similar to \eqref{eq:topic_embedding}, we find the $M$ words with the closest embeddings to each cluster center to represent the topic and compute the topic embedding $\underline{t}_j$ as the weighted average embedding of $M$ words in the $j$th topic. Among all $J$ cluster centers, we pick the $K$ topics with the closest $\underline{t}_j$ to the prompt embedding, where the prompt embedding is the average embedding of all words in the input prompt.


The second type of methods discovers the $K$ topics from the input prompt. We cluster non-stop words in the prompt using non-negative sparse coding~\cite{hoyer2002non} (\textbf{NNSC-local}) and Kmeans (\textbf{Kmeans-local}). 
We also sample $K$ non-stop words from the prompt and call it \textbf{Sample-local}. Similar to \eqref{eq:topic_embedding}, we represent each topic using $M$ words and compute the weighted average of their embeddings $\underline{t}_k$ as the input of our text generator. Notice that the locally clustering methods produce similar results when the prompts come from STSb due to their short lengths, so we only test \textbf{Kmeans-local} in our human evaluation.










\subsubsection{Results}
In Table~\ref{tb:option_auto}, we show that local methods generate the options more relevant to the input prompt than the global methods due to significantly higher Sim and Sim Short. Our method performs better compared to other local methods, especially in Sim Diff, which highlights the high novelty of our generated topics. The improvement on Sim Short is larger than that on Sim because our method could suggest the related topics that are not explicitly mentioned in the short prompt (e.g., \emph{U.S.} in Figure~\ref{fig:first_page}).

The human evaluation results are presented in Table~\ref{tb:option_human}. Our method wins in terms of generating relevant topics that promote the narrative. The \textbf{Kmeans-local} performs better in L because most of the words in the input prompts could be mentioned again in the next sentence. However, it often leads to the redundant topics that are too similar to the prompt. 

Table~\ref{tb:generated_topics} compares the options generated by different methods while Table~\ref{tb:generated_text_examples} compares the text generated using different option generators and text generators. More examples are presented in \refAppendixMoreExample. In Table~\ref{tb:generated_topics}, we can see that most topics in \textbf{Kmeans-local} do not promote the narrative, which makes the generated continuation become a copy of the input prompt in Table~\ref{tb:generated_text_examples}. We will quantitatively evaluate the generated continuations using different option generators in \refAppendixMoreExp. Notice that the high redundancy problem is hard to be solved by a conditional text generator because the relatedness between the prompt and the generated text is hard to be controlled~\cite{see2019makes}.









\begin{table*}[t!]
\scalebox{0.8}{
\begin{tabular}{|c|ccc|ccc|c|cc|c|}
\hline
Text & \multicolumn{6}{|c|}{Automatic Metrics} & Inference & \multicolumn{3}{|c|}{Human Judgement} \\ \cline{2-7} \cline{9-11}
Generation & \multicolumn{3}{|c|}{Relevancy Hit} & \multicolumn{3}{|c|}{Quality} & Time & \multicolumn{2}{|c|}{Relevancy} & Fluency \\ \cline{2-11}

Method &  Token & Word & Topic & PPL ($\downarrow$) & Dist-1 & Dist-2  & s ($\downarrow$) & Recall & Precision & Score \\  \hline
PPLM & 1.48 & 0.99 & 0.77 & 18.49 &  \textbf{40.29} & \textbf{80.83} & 17.74 & 30.56 $\pm$ 2.96 & 56.01 $\pm$ 4.41 & 3.83 $\pm$ 0.13 \\
Ours & \textbf{2.36} & \textbf{1.79} & \textbf{1.40} & \textbf{16.39} & 37.98 & 79.65 & \textbf{1.02} &  \textbf{41.46} $\pm$ 3.47 & \textbf{56.41} $\pm$ 4.41 & \textbf{4.07} $\pm$ 0.10 \\ \hline \hline
GPT2 & 1.27 & 0.84 & 0.64 & 14.24 & 39.80 & 80.22 & 1.00 & 24.49 $\pm$ 2.77 & 48.69 $\pm$ 4.61 & 4.15 $\pm$ 0.11 \\ \hline 


\end{tabular}
}
\centering
\caption{Comparison of conditional text generators. The numbers in Dist-1, Dist-2, Recall, and Precision are percentages. Lower perplexity (PPL) and inference time are better. The better performances between PPLM and our method are highlighted. In human evaluation, we report the mean $\pm$ standard error of each method.}
\label{tb:generation_auto}
\end{table*}

\subsection{Conditional Text Generator  Evaluation}

To demonstrate our text generator's effectiveness, we use our option generator to prepare the topic embeddings and randomly select $n$ topics as our conditions to simulate the user's choice, where $n$ is a random number from 1 to $K$. The sentences generated by different methods are compared.





\subsubsection{Automatic Evaluation Metrics}
We match the union of $M \times K$ top words in the chosen topics with the words in the generated continuations and count the number of tokens that are matched exactly (token), the number of matched word types (word), and the number of topics that contain at least one matched word (topic) to measure the relevancy between the continuations and the chosen topics. Notice that the scores are underestimated because the generation might mention words in different morphological variations or other words related to the topics. 



The fluency of the generated text is measured using the perplexity~\cite{serban2016building} of the original GPT2 (with 345M parameters) without being fine-tuned on Wikipedia. Dist-n~\cite{li2016diversity} is the ratio between the number of unique n-grams and the number of all n-grams in the continuations, where n=1 or 2. Higher Dist-n implies more diverse generations.
The average inference time per input prompt is also presented.



\subsubsection{Human Evaluation}
We present the prompt and the generated continuation and ask the worker to score the generation's fluency from 1 (not fluent at all) to 5 (very fluent). Next, we show $K$ topics and ask which topics are mentioned in the generation. Treating the worker's choices as prediction and the topics our model conditions on as ground truth, we report the average precision and recall of the prediction.


\subsubsection{Conditional Text Generator Baselines}
We compare our method with \textbf{PPLM} (Plug and Play Language Models)~\citep{dathathri2019plug} due to its strong performance against the weighted decoding approach from~\citet{ghazvininejad2017hafez} when the condition is a bag of words. 

The condition for \textbf{PPLM} is the union of the top $M$ words in the chosen topics and each word's weight is neglected. We use our generation model without conditioning on any word (i.e., $n=0$) during testing\footnote{We find the model performs similarly compared with the GPT2 with no condition during training.} as the base model of \textbf{PPLM}.
We also present the performance of the base model itself as a reference to know the significance of our improvement (denoted as \textbf{GPT2}). 



\subsubsection{Results}

Table~\ref{tb:generation_auto} indicates that our model outperforms \textbf{PPLM} in all metrics except in Dist-1 and Dist-2. We suspect that our model generates slightly less diverse sentences in order to make the generation more relevant to the given topics.  


The generation might mention a topic even if it is not chosen as a condition, so we achieve similar precision compared to \textbf{PPLM} in human evaluation. The recall of \textbf{PPLM} means that only around 30\% of given topics are mentioned. The low recall indicates the difficulty of mentioning multiple randomly selected topics in the next 50 word pieces while keeping the sentence fluent. By contrast, achieving 40\% on recall demonstrates the effectiveness of our conditional text generator.




Compared with \textbf{PPLM}, our model requires an additional training step but achieves low inference time and high relevancy to the given topics/words once the training is finished. The benefits make it preferable in our interactive writing application. 







\section{Related Work}

Different interactive writing assistants provide different forms of options to let users express their preferences. The options could be manually defined classes (e.g., sentiment)~\cite{keskar2019ctrl, dathathri2019plug}, semantic frames~\cite{tu2019generating}, or event structures such as (subject, verb, object, modifier)~\cite{martin2018event,tambwekar2019controllable, ammanabrolu2019story}. The forms of options allow users to control the attributes of the generated text but require labels or classifiers that map the text to the attributes/options.

The options could also be a single query word at the beginning~\cite{austin2019book}, the article title~\cite{yan2016poet}, politeness~\cite{niu2018polite} or specificity~\cite{see2019makes} of the text, or the length of the generated sentence~\cite{tu2019generating}. However, the options cannot provide fine-grained control on topical directions of the generated contents.

A related research direction is the multi-stage story generation. To make a long story more coherent, recent work proposes to generate a skeleton and then generate the full text guided by the skeleton. The skeleton could be a sequence of SRL frames~\cite{fan2019strategies}, a sequence of event structure (subject, verb, object, preposition, modifier)~\cite{ammanabrolu2019story}, a story premise~\cite{fan2018hierarchical}, or a story summary~\cite{chen2019learning}. Users can revise the skeleton to control the generated text, but the approaches assume the existence of the skeleton extractor or labels in the training corpus.
Besides, the systems cannot suggest options given the partial text, which is one of the main focuses of our interactive writing assistant.

The skeleton could also be multiple keyphrases. The keyphrases are extracted based on word frequency~\cite{ippolito2019unsupervised, tan2020progressive, wu2020controllable}, an off-the-shelf keyword extraction method~\cite{peng2018towards, goldfarb2019plan, yao2019plan, rashkin2020plotmachines, zhang2020pointer}, a sentence compression dataset and reinforcement learning~\cite{xu2018skeleton}, or image caption datasets and ConceptNet~\citep{lin2020commongen}. Most of the studies focus on modeling the long-term dependency among the keyphrases and/or forcing the generation to contain the keyphrases. Instead, we focus on allowing users to determine the topical directions of the generation. Compared with conditioning on keyphrases, our interactive writing assistant is especially helpful when users do not know the exact phrases they want to see or when the given keyphrase extractor does not detect the desired topics.

\section{Conclusion}
We propose an interactive writing assistant that generates topic options given an input prompt and generates the continuation of the prompt given the topics chosen by a user. We decompose the framework into two components and propose a novel model for each component. The automated evaluation and human evaluation indicate that our system generates many topics that are related to but different from the prompt, and generates the sentences that are fluent and relevant to the chosen topics. 

\section*{Acknowledgements}
We thank Ao Liu for his preliminary exploration of this project and Nader Akoury for his helpful feedbacks. We also thank the anonymous reviewers for their constructive feedback.

This work was supported in part by the Center for Data Science and the Center for Intelligent Information Retrieval, in part by the Chan Zuckerberg Initiative under the project Scientific Knowledge Base Construction, in part using high performance computing equipment obtained under a grant from the Collaborative R\&D Fund managed by the Massachusetts Technology Collaborative, in part by the National Science Foundation (NSF) grant numbers DMR-1534431 and IIS-1514053. 

Any opinions, findings, conclusions, or recommendations expressed in this material are those of the authors and do not necessarily reflect those of the sponsor.

\bibliography{ref}
\bibliographystyle{acl_natbib}

\newpage
\appendix

\section{Option Generator Comparison Using Generated Continuations}
\label{sec:more_experiment}

\begin{table}[t!]
\scalebox{0.75}{
\begin{tabular}{|c|c|ccc|}
\hline
Scope & Method & F & NP & A \\ \hline
\multirow{2}{*}{Global} & LDA & 3.07 $\pm$ 0.17 & 2.82 $\pm$ 0.16 & 3.06 $\pm$ 0.13 \\ 
 & Kmeans & 3.65 $\pm$ 0.13 & 3.42 $\pm$ 0.14 & 3.42 $\pm$ 0.12 \\ \hline
\multirow{2}{*}{Local} & Kmeans & 3.71 $\pm$ 0.13 & 3.56 $\pm$ 0.15 & 3.39 $\pm$ 0.13 \\
& Ours & \textbf{3.85} $\pm$ 0.14 & \textbf{3.64} $\pm$ 0.15 & \textbf{3.67} $\pm$ 0.14 \\ \hline

\end{tabular}
}
\centering
\caption{ Comparison of the continuations generated by different option generators using human judgment (mean $\pm$ standard error). F, NP, and A refer to fluency, narrative promotion, and overall, respectively.}
\label{tb:option_human_2}
\end{table}

\begin{table*}[t!]
\scalebox{0.85}{
\begin{tabular}{|c|c|cccccc|}
\hline
Scope & Method &  BLEU & BLEU Diff & Word Hit & Self-BLEU ($\downarrow$) & Dist-1 & Dist-2  \\  \hline

\multirow{3}{*}{Global} & Sample &  \textbf{7.39} &\textbf{5.66} & 0.34 & \textbf{9.45} & \textbf{47.60} & \textbf{86.79} \\
& LDA  & 7.19 & 4.87 & \textbf{2.01} & 13.06 & 36.02 & 78.73  \\
& Kmeans & 7.12 & 4.65 & 1.30 & 12.23 & 36.62 & 81.49 \\ \hline
\multirow{4}{*}{Local} & Sample & 8.38 & 2.71 & 2.93 & 18.03 & 35.76 & 77.00 \\
& NNSC & \textbf{8.44} & 3.24 & 2.94 & 17.20 & 35.43 & 76.71 \\
& Kmeans & 8.32 & 3.06 & 2.96 & 16.97 & 35.39 & 77.10 \\
& Ours & 8.38 & \textbf{5.55} & \textbf{3.02} & \textbf{15.97} & \textbf{36.18} & \textbf{78.71}  \\ \hline \hline
NA & None & 8.50 & 5.59 & - & 13.17 & 39.69 & 80.17 \\ \hline
\end{tabular}
}
\centering
\caption{Comparison of the continuations generated by different option generators using automatic metrics. The values are percentages except in Word Hit. Higher numbers are better except in Self-BLEU. The best numbers within each scope are highlighted.}
\label{tb:option_auto_2}
\end{table*}

To see whether the proposed option generator improves the quality of the continuations, we use all of $K$ topics from different methods to guide our conditional text generator and compare their generated continuations. In addition to all the methods we described in Section \refoptgenmethod, we also present the results of our text generator without conditioning on any topics (i.e., $n=0$) as a reference and call the method \textbf{None}.

\subsection{Automatic Evaluation Metrics}
\begin{itemize}
\setlength\itemsep{-0.3em}
\item \textbf{BLEU}: 
For each generated text guided by the set of $K$ topics, we report BLEU-2~\citep{papineni2002bleu} between the generation and the actual continuation containing $O=25$ words. We adopt the smoothing method 3 in~\citet{chen2014systematic} because there is sometimes no bigram overlapping between the predicted continuation and the actual continuation.


\item \textbf{BLEU Diff}: 
Similar to Sim Diff, BLEU Diff is the BLEU score between the generation and the continuation minus the BLEU score between the generation and the input prompt.

\item \textbf{Word Hit}: 
If the generated topics are not relevant to the input prompt, our conditional text generator might have difficulty in mentioning the related words in the continuation. We report how many unique word types representing $K$ topics are mentioned in the generated continuation.

\item \textbf{Self-BLEU}: The metric computes the average pairwise BLEU scores of 3 generations~\cite{zhu2018texygen}. Lower Self-BLEU implies the options encourage more diverse generations.


\end{itemize}

\subsection{Human Evaluation}
We show the continuation guided by all topics and ask how fluent the sentence is (F), how helpful the sentence can promote the narrative (NP), and the overall quality of the generation (A). The worker can choose from 5 options, and 5 means very fluent, very helpful, and excellent, respectively.

\subsection{Results}
The automatic evaluation results are presented in Table~\ref{tb:option_auto_2}. As expected, the options generated by the local methods lead to the continuations that are more similar to the actual continuation (i.e., higher BLEU score) compared to that generated by the global methods. Global topics encourage the generated text to be unrelated to the input prompt, so leading to more diverse sentences (i.e., lower Self-BLEU and higher Dist-1 and Dist-2).

Our method performs better in most metrics than the other local methods, especially in BLEU Diff, while achieving comparable BLEU, which means our generated options often result in the relevant and diverse continuations that are sufficiently different from the prompt. Furthermore, the human evaluation results in Table~\ref{tb:option_human_2} show that our method outperforms other baselines in all metrics.

\section{Implementation Details}
\label{sec:implement_details}

The training algorithm for our option generator could be seen in Algorithm~\ref{train_algo}. The algorithm is similar to the training method in~\citet{NSD}. For each non-stop word in the continuation $\bar{y}_o$, we linearly combine all the cluster centers $\underline{c}_1,...\underline{c}_K$ to reconstruct the word embedding of $\bar{y}_o$. We only allow positive weights, $a_1,...,a_K \geq 0$, and incorporate L1 loss $\sum\limits_{k=1}^K a_k$ to encourage the weights of the irrelevant cluster centers to be 0, so the clustering method is called non-negative sparse coding (NNSC)~\citep{hoyer2002non}.  Estimating $a_1,...,a_K$ could be viewed as E-step, which matches the clusters and the word embedding in the continuation. In the M-step, we fix the estimated weights $\hat{a}_1,...,\hat{a}_K$ and use backpropagation to encourage the cluster centers to be closer to the embedding of $\bar{y}_o$. To encourage the cluster centers to be context dependent, we also use the same EM optimization to push away the clusters centers from negative samples' embeddings.

During training, the input prompt is tokenized into word pieces, and the actual continuation is tokenized into words. We run the byte pair encoding~\citep{sennrich2015neural} to get word pieces required by GPT2 and run Spacy tokenizer\footnote{\url{spacy.io/}} to get words required by GloVe. The two tokenization results are aligned to collect the training examples.

\begin{algorithm*}[!t]
\SetAlgoLined
\SetKwInOut{Input}{Input}
\SetKwInOut{Output}{Output}

\Input{Training corpus, stop word list, pretrained GPT2 encoder, and pre-trained word embeddings.}
\Output{Neural option generator}
Initialize our encoder using a pretrained GPT2 model and randomly initialize the other parameters\\
\ForEach{$x_1,...,x_I$ \text{in training corpus}}{%
	Run forward pass of our model given $x_1,...,x_I$ to compute the cluster centers $\underline{c}_1,...\underline{c}_K$ \\
	
	Collect the positive examples $\bar{y}_1,...,\bar{y}_{O}$ (i.e., non-stop words after $x_I$) and their word embeddings $\underline{e}_o^{\bar{y}}$ \\
	Collect the negative examples $\bar{y}'_1,...,\bar{y}'_{O}$ (i.e., a randomly sampled continuation without stop words) and their word embeddings $\underline{e}_o^{\bar{y}'}$ \\
	$L = 0$ \\
	\ForEach{$\bar{y}_o$ \text{in the positive example}}{
	    Estimate $\hat{a}_1,...,\hat{a}_K = \argmin\limits_{0 \leq a_1,...,a_K \leq 1} ||\sum\limits_{k=1}^K a_k \underline{c}_k - \underline{e}_o^{\bar{y}}||^2 + \lambda \sum\limits_{k=1}^K a_k$ using RMSprop \\
	    $L = L + ||\sum\limits_{k=1}^K \hat{a}_k \underline{c}_k - \underline{e}_o^{\bar{y}}||^2$
	}
	\ForEach{$\bar{y}'_o$  \text{in the negative example}}{
	    Estimate $\hat{b}_1,...,\hat{b}_K = \argmin\limits_{0 \leq b_1,...,b_K \leq 1} ||\sum\limits_{k=1}^K b_k \underline{c}_k - \underline{e}_o^{\bar{y}'}||^2 + \lambda \sum\limits_{k=1}^K b_k$ using RMSprop\\
	    $L = L - ||\sum\limits_{k=1}^K \hat{b}_k \underline{c}_k - \underline{e}_o^{\bar{y}'}||^2$
	}
    Update our neural model by backpropagation through cluster centers $\underline{c}_1,...\underline{c}_K$ to minimize $L$
}
 \caption{Training procedure for our option generator (using batch size = 1)}
 \label{train_algo}
\end{algorithm*}

When training our option generator, we sample a word piece sequence with length 512 as the input of the GPT2 encoder. We randomly select a number from 1 to 199 as the size of the first input prompt and the next prompt always contains 200 more word pieces than the previous one. Each continuation includes 50 words (not including stop words) after the corresponding prompt. In the same text sequence, the last output embedding in every prompt receives gradients together from a single backward pass. We initialize our encoder using distilled GPT2~\citep{sanh2019distilbert} to save GPU memory and the parameters are trained using SGD as in~\citet{NSD}. 

When training our conditional text generator, the size of the input to the GPT2 encoder is 256. We randomly select 5 positions from the input sequence to insert the future words sampled from the continuation containing 25 words (after removing stop words). Although we insert future words into multiple positions to speed up the training, we insert the future words once (only before the end of the prompt) during testing. We initialize our encoder using the GPT2 with 117M parameters and train the parameters using AdamW~\citep{loshchilov2017decoupled}. Notice that we insert at most $K$ words before each position during training. Therefore, the number of specified words plus the number of chosen topics cannot be greater than $K$ during testing.

We use the cased version (840B) of GloVe embedding. 
The GloVe embedding in both components is fixed to allow the two components that are trained parallelly to communicate during testing. To simplify our method, we train the two components separately and bridge the components using GloVe.\footnote{If we want to let the text generator directly condition on the topics rather than words during training, we need to know what topics that are mentioned by the actual continuation and how often our option generator predicts the topics. Trying the achieve this will complicate the method, so we leave this direction as future work.} 
Training separately also allows the language generator to use a larger model on a GPU with limited memory. We use a GTX TITAN X and train the option generator for around three weeks and train the conditional text generator for about five weeks.





\begin{table*}[t!]
\scalebox{0.45}{
\begin{tabular}{|cccc|cccc|cccc|}
\hline
\multicolumn{2}{|c|}{ {\Large \Gape[3pt][1pt]{Input Prompt}} } &  \multicolumn{10}{|c|}{ {\Large defense chiefs from estonia, latvia, lithuania, germany, italy, spain and slovakia signed the agreement.} }  \\ \hline
\multicolumn{4}{|c|}{ {\Large \Gape[3pt][1pt]{LDA-global} } } & \multicolumn{4}{|c|}{ {\Large Kmeans-local} } & \multicolumn{4}{|c|}{ {\Large Ours} } \\ \hline
1&police, prison&6&Draft, NCAA&1&defense, defenses&6&signed, signing&1&century, Roman&6&1754, 1744 \\ 
2&football, basketball&7&League, league&2&chiefs, chieftains&7&defense, defenses&2&constitutional, mandate&7&knew, wished \\ 
3&Nations, Foreign&8&company, subsidiary&3&signed, signing&8&agreement, agreements&3&king, prince&8&Bulgars, Magyars \\ 
4&company, companies&9&baseball, Baseball&4&agreement, agreements&9&signed, signing&4&Romanian, Hungarian&9&troops, war \\ 
5&party, Party&10&game, games&5&chiefs, chieftains&10&defense, defenses&5&kingdom, kings&10&Slovakia, Latvia \\ 
\hline \hline 
\multicolumn{2}{|c|}{ {\Large \Gape[3pt][1pt]{Input Prompt}} } &  \multicolumn{10}{|c|}{ {\Large The two Democrats on the five-member FCC panel held a news conference to sway opinion against Powell.} }  \\ \hline
\multicolumn{4}{|c|}{ {\Large \Gape[3pt][1pt]{LDA-global} } } & \multicolumn{4}{|c|}{ {\Large Kmeans-local} } & \multicolumn{4}{|c|}{ {\Large Ours} } \\ \hline
1&Republican, Democratic&6&company, companies&1&conference, conferences&6&Democrats, Republicans&1&CNN, news&6&said, stated \\ 
2&party, Party&7&psychology, research&2&news, headlines&7&member, held&2&Committee, Legislative&7&know, sure \\ 
3&election, elections&8&football, basketball&3&panel, panels&8&opinion, opinions&3&party, Party&8&culminated, protested \\ 
4&television, show&9&Nations, Foreign&4&FCC, CRTC&9&sway, sways&4&Smith, Thompson&9&election, ballot \\ 
5&police, prison&10&James, Robert&5&Powell, Thompson&10&three, four&5&telecommunications, corporations&10&Obama, Barack \\ 
\hline \hline 
\multicolumn{2}{|c|}{ {\Large \Gape[3pt][1pt]{Input Prompt}} } &  \multicolumn{10}{|c|}{ {\Large The MSN Messenger 6 software will be available from 11 a.m. PST on Wednesday, according to Microsoft.} }  \\ \hline
\multicolumn{4}{|c|}{ {\Large \Gape[3pt][1pt]{LDA-global} } } & \multicolumn{4}{|c|}{ {\Large Kmeans-local} } & \multicolumn{4}{|c|}{ {\Large Ours} } \\ \hline
1&software, user&6&California, Disney&1&PST, PDT&6&Messenger, messenger&1&integration, development&6&2012, February \\ 
2&company, companies&7&cards, dog&2&a.m., p.m.&7&9, 6&2&configuration, interface&7&provide, available \\ 
3&television, show&8&company, subsidiary&3&available, Available&8&according, According&3&websites, web&8&6, 9 \\ 
4&game, games&9&party, Party&4&will, must&9&Wednesday, Tuesday&4&released, release&9&IPv6, InfiniBand \\ 
5&Education, College&10&radio, FM&5&software, Microsoft&10&MSN, Yahoo&5&smartphones, smartphone&10&Windows, Desktop \\ 
\hline \hline 
\multicolumn{2}{|c|}{ {\Large \Gape[3pt][1pt]{Input Prompt}} } &  \multicolumn{10}{|c|}{ {\Large Schools that fail to meet state goals for three years in a row must offer tutoring in addition to transfers.} }  \\ \hline
\multicolumn{4}{|c|}{ {\Large \Gape[3pt][1pt]{LDA-global} } } & \multicolumn{4}{|c|}{ {\Large Kmeans-local} } & \multicolumn{4}{|c|}{ {\Large Ours} } \\ \hline
1&company, companies&6&software, user&1&must, meet&6&fail, failing&1&learning, concepts&6&funded, nonprofit \\ 
2&football, basketball&7&cards, dog&2&row, rows&7&years, year&2&Education, Curriculum&7&need, able \\ 
3&psychology, research&8&patients, treatment&3&three, four&8&tutoring, tutor&3&students, student&8&five, six \\ 
4&game, games&9&police, prison&4&goals, goal&9&transfers, offer&4&applicant, stipulated&9&tax, taxes \\ 
5&Education, College&10&population, households&5&addition, additional&10&Schools, School&5&school, kindergarten&10&State, Missouri \\ 
\hline \hline 
\multicolumn{2}{|c|}{ {\Large \Gape[3pt][1pt]{Input Prompt}} } &  \multicolumn{10}{|c|}{ {\Large Declining issues outnumbered advancers slightly more than 3 to 1 on the New York Stock Exchange.} }  \\ \hline
\multicolumn{4}{|c|}{ {\Large \Gape[3pt][1pt]{LDA-global} } } & \multicolumn{4}{|c|}{ {\Large Kmeans-local} } & \multicolumn{4}{|c|}{ {\Large Ours} } \\ \hline
1&County, Historic&6&Rhode, Connecticut&1&New, York&6&Stock, stock&1&economic, economy&6&1848, 1859 \\ 
2&California, Disney&7&Angeles, Los&2&1, 2&7&York, NY&2&Investment, Financial&7&even, enough \\ 
3&Canada, Ontario&8&Australian, Melbourne&3&3, 4&8&Declining, Decline&3&bank, loans&8&4, 3 \\ 
4&company, companies&9&Nations, Foreign&4&Exchange, exchange&9&outnumbered, outnumbering&4&$, US$&9&\%, percent \\ 
5&China, Hong&10&Education, College&5&issues, issue&10&slightly, somewhat&5&market, trading&10&York, New \\ 
\hline \hline 
\multicolumn{2}{|c|}{ {\Large \Gape[3pt][1pt]{Input Prompt}} } &  \multicolumn{10}{|c|}{ {\Large The Portuguese weather service said Europe's heatwave was caused by a mass of hot, dry air moving from the southeast.} }  \\ \hline
\multicolumn{4}{|c|}{ {\Large \Gape[3pt][1pt]{LDA-global} } } & \multicolumn{4}{|c|}{ {\Large Kmeans-local} } & \multicolumn{4}{|c|}{ {\Large Ours} } \\ \hline
1&chemical, carbon&6&police, prison&1&hot, sexy&6&northeast, weather&1&population, estimates&6&October, February \\ 
2&company, companies&7&island, Island&2&mass, masses&7&caused, causing&2&temperature, heat&7&seemed, just \\ 
3&park, Park&8&restaurant, food&3&heatwave, downpours&8&air, Air&3&storm, storms&8&35, 10 \\ 
4&plant, plants&9&River, river&4&Europe, European&9&dry, drying&4&Pedro, Vicente&9&caused, severe \\ 
5&engine, aircraft&10&brown, grey&5&moving, said&10&Portuguese, Spanish&5&north, south&10&Portugal, Spain \\ 
\hline \hline 
\multicolumn{2}{|c|}{ {\Large \Gape[3pt][1pt]{Input Prompt}} } &  \multicolumn{10}{|c|}{ {\Large tibet suspects chinese government of creating the virus to spy on tibetan exiles and the dalai lama.} }  \\ \hline
\multicolumn{4}{|c|}{ {\Large \Gape[3pt][1pt]{LDA-global} } } & \multicolumn{4}{|c|}{ {\Large Kmeans-local} } & \multicolumn{4}{|c|}{ {\Large Ours} } \\ \hline
1&police, prison&6&story, book&1&suspects, suspect&6&spy, spies&1&film, movie&6&tells, asks \\ 
2&African, Africans&7&Iranian, Iran&2&chinese, japanese&7&exiles, exile&2&government, governmental&7&want, know \\ 
3&psychology, research&8&Nations, Foreign&3&government, governments&8&lama, Lama&3&military, government&8&insurrectionists, reactionaries \\ 
4&software, user&9&party, Party&4&creating, create&9&creating, create&4&Lai, Ying&9&killed, killing \\ 
5&cards, dog&10&China, Hong&5&virus, viruses&10&lama, Lama&5&creatures, creature&10&Thailand, Malaysia \\ 
\hline \hline 
\multicolumn{2}{|c|}{ {\Large \Gape[3pt][1pt]{Input Prompt}} } &  \multicolumn{10}{|c|}{ {\Large I have  years of "Neener Neener" rights Usually I get pretty decent care.} }  \\ \hline
\multicolumn{4}{|c|}{ {\Large \Gape[3pt][1pt]{LDA-global} } } & \multicolumn{4}{|c|}{ {\Large Kmeans-local} } & \multicolumn{4}{|c|}{ {\Large Ours} } \\ \hline
1&cards, dog&6&psychology, research&1&years, year&6&decent, good&1&film, films&6&said, told \\ 
2&company, companies&7&television, show&2&rights, Rights&7&care, health&2&song, lyrics&7&really, know \\ 
3&story, book&8&software, user&3&Usually, Normally&8&years, year&3&album, albums&8&downright, cynical \\ 
4&game, games&9&football, basketball&4&get, getting&9&get, getting&4&Sommer, Steffen&9&expressive, portrayal \\ 
5&patients, treatment&10&African, Africans&5&pretty, quite&10&get, getting&5&girl, teenage&10&Germany, Berlin \\ 
\hline \hline 
\end{tabular}
}
\centering
\caption{Comparison of all $K$ topics for the input prompts using $M=2$ words closest to each topic. }
\label{tb:generated_more_topics}
\end{table*}

\begin{table*}[t!]
\scalebox{0.68}{
\begin{tabular}{|cc|c|}
\hline
\multicolumn{2}{|c|}{Input Prompt} &  defense chiefs from estonia, latvia, lithuania, germany, italy, spain and slovakia signed the agreement. \\ \hline
\multicolumn{2}{|c|}{Generator} & \multirow{2}{*}{Generated Text}  \\ \cline{1-2} 
 Option & Text &   \\ \hline
LDA-global & Ours &  After the talks, the League of Nations allowed the German Democratic Republic's representatives to negotiate the deal.  \\
Kmeans-local & Ours &  These agreements were based on the agreement signed by the German king Frederick Barbarossa between 870 and 873 \\
Ours & PPLM &  For the period of five years in Lithuania were the chief ministers (procurator princeps or jevgadirs) and the chief  \\
None & GPT2 &  ( This treaty would come under Royal Decree 1282 on 8 September 1725.) On 9 December 1725, Russian armies entered  \\
Ours & Ours &  These agreements were signed in 1756 by the sovereigns of Moldavia (Moorish) and the princely states of the Romanian  \\ \hline \hline
\multicolumn{2}{|c|}{Input Prompt} &  The two Democrats on the five-member FCC panel held a news conference to sway opinion against Powell. \\ \hline
\multicolumn{2}{|c|}{Generator} & \multirow{2}{*}{Generated Text}  \\ \cline{1-2} 
 Option & Text &   \\ \hline
LDA-global & Ours &  He and his former friend and fellow Democrat, James H.  Jim  White, were arrested on charges of corruption and child  \\
Kmeans-local & Ours &  He and his three other Democrats had no time to discuss the other four; this changed at the conference and at the FCC \\
Ours & PPLM &  She responded,  The House has decided, 'When the other candidates say, 'Let Democrats take over the FCC,' it's kind of  \\
None & GPT2 &  When questioned in the news, Powell stated  The fact that she does not want to get a job with a group that includes me  \\
Ours & Ours &  As a result, a Senate committee investigation  by the Senate  said that the Democratic party had been involved in the \\ \hline \hline
\multicolumn{2}{|c|}{Input Prompt} &  The MSN Messenger 6 software will be available from 11 a.m. PST on Wednesday, according to Microsoft. \\ \hline
\multicolumn{2}{|c|}{Generator} & \multirow{2}{*}{Generated Text}  \\ \cline{1-2} 
 Option & Text &   \\ \hline
LDA-global & Ours &  Thessaloniki Business Center - a business school that was built in 1995 and that provides jobs in business, technology, \\
Kmeans-local & Ours &  The MSN Messenger 8 software will be available from 9 a.m. PST on Thursday, according to Microsoft. The MSN \\
Ours & PPLM &  On Tuesday January 22, 2016, Microsoft announced that the Internet Mail service, the Messenger Plus service, is going  \\
None & GPT2 &  Microsoft plans to expand the coverage of MSN Messenger in the United States.. nbc.org; November 8, 2008. In its  \\
Ours & Ours &  The Windows Messenger 6 web app now has a new web service   for mobile devices to download the Windows product.  \\ \hline \hline
\multicolumn{2}{|c|}{Input Prompt} &  Schools that fail to meet state goals for three years in a row must offer tutoring in addition to transfers. \\ \hline
\multicolumn{2}{|c|}{Generator} & \multirow{2}{*}{Generated Text}  \\ \cline{1-2} 
 Option & Text &   \\ \hline
LDA-global & Ours &  Students at the School of Business, Computer, and information science programs must complete their coursework in  \\
Kmeans-local & Ours &  Tuition on non-residential loans, and the availability of tutoring for at least three years, will be phased out before  \\
Ours & PPLM &  The school also provides scholarships to students from the other districts who apply for the school to receive free or  \\
None & GPT2 &  The program is an outgrowth of the Tisch School's efforts to build the academic program required for graduate programs, \\
Ours & Ours &  Additionally, it must also provide a forum to discuss the learning needs of its students. California school districts a \\ \hline \hline
\multicolumn{2}{|c|}{Input Prompt} &  Declining issues outnumbered advancers slightly more than 3 to 1 on the New York Stock Exchange. \\ \hline
\multicolumn{2}{|c|}{Generator} & \multirow{2}{*}{Generated Text}  \\ \cline{1-2} 
 Option & Text &   \\ \hline
LDA-global & Ours &  New York City      San Mateo County      Windsor County, Ontario       Windsor, Ontario (convention)        Windsor, On \\
Kmeans-local & Ours &  New York Stock Exchange had been down for issues outnumbered 7 to 1 on the New York Exchange. Declining issues  \\
Ours & PPLM & , Financial Times   According to Bloomberg's Financial Times,  the firm's current capital flows to the Securities    \\
None & GPT2 &  On December 19, 2006,    and February 10, 2007, respectively,    and February 21, 2007, respectively,    and February 5 \\
Ours & Ours &  In response the Bank of New York announced \$100 million in loan interest. The Bank withdrew its \$5 million offer,  \\ \hline \hline
\multicolumn{2}{|c|}{Input Prompt} &  The Portuguese weather service said Europe's heatwave was caused by a mass of hot, dry air moving from the southeast. \\ \hline
\multicolumn{2}{|c|}{Generator} & \multirow{2}{*}{Generated Text}  \\ \cline{1-2} 
 Option & Text &   \\ \hline
LDA-global & Ours &  The police said it is likely that the heat wave is coming from the ocean around the park. There are a number of other  \\
Kmeans-local & Ours &  The North American weather service said Europe's heatwave was caused by a mass of hot, dry air moving from the  \\
Ours & PPLM &  It was a record in the European part of the Western Hemisphere. At 2AM Eastern Europe will see two nights a week of  \\
None & GPT2 &  On January 1, 2014, the station's digital channel was shut down as digital television began broadcasting, ending  \\
Ours & Ours &  The most common weather to affect Portugal was the August, which began with an approaching hurricane, causing severe \\ \hline \hline
\multicolumn{2}{|c|}{Input Prompt} &  tibet suspects chinese government of creating the virus to spy on tibetan exiles and the dalai lama. \\ \hline
\multicolumn{2}{|c|}{Generator} & \multirow{2}{*}{Generated Text}  \\ \cline{1-2} 
 Option & Text &   \\ \hline
LDA-global & Ours &  In 2003, Chinese Nationalist Party leader Deng Xiaoping visited Taiwan and told the Chinese Nationalist Party  \\
Kmeans-local & Ours &  He warns Tulku and Chinkua against spreading it because it is a secret operation by the Chinese and is a spy on the  \\
Ours & PPLM &  They then begin to research the chinese hukou and chinta. They find that the Chinese believe that they are agents of  \\
None & GPT2 &  Although not suspected, one person known of the virus to be infected with it in Thailand is unknown. The virus  \\
Ours & Ours &  He threatens to reveal his plans with the aid of Malaysia government with all the help of Japan and China to the  \\ \hline \hline
\multicolumn{2}{|c|}{Input Prompt} &  I have  years of "Neener Neener" rights Usually I get pretty decent care. \\ \hline
\multicolumn{2}{|c|}{Generator} & \multirow{2}{*}{Generated Text}  \\ \cline{1-2} 
 Option & Text &   \\ \hline
LDA-global & Ours &  A computer games system or computer science is a game system that is designed in a way that a computer is capable of  \\
Kmeans-local & Ours &  In later years, my patients never get very good care. The rights are only a couple to a day so I've just had a few year \\
Ours & PPLM &  I hope you can bring in more people. We've a good relationship. We've got a good relationship with our customers since  \\
None & GPT2 &  I've no idea how much care I get with the NERVIC\textunderscore CORE\textunderscore GILAS\textunderscore WELL\textunderscore CORE\textunderscore GILAS\textunderscore WELL\textunderscore CORPS\textunderscore VIGILUS) \\
Ours & Ours &  The album 'Nederlands Kort   Eindhoven,   which, when I mentioned, was a pop-rock film, was downright cynical - a song  \\ \hline \hline
\end{tabular}
}
\centering
\caption{The continuations that are generated by conditioning on all of $K$ topics from different option generators. The input prompts comes from STSb.
}
\label{tb:generated_more_text_examples}
\end{table*}

\section{Experiment Details}
\label{sec:experiment_details}

We truncate the probabilities after the top 40 in top-k sampling~\citep{fan2018hierarchical}. In all the experiments, we set $M = 5$ words to represent each topic, although the figures and tables use $M = 2$ or $M = 3$ due to the space limit. We set $K=10$ because $K=10$ seems to work well in~\citet{NSD}. Our Transformer decoder for option generation has 5 layers. 

In the following subsections, we describe the details about our baselines, the automatic evaluation, and human evaluation.

\subsection{Baselines}

We adopt the default hyper-parameters of LDA in gensim\footnote{\url{ https://radimrehurek.com/gensim/models/ldamulticore.html}}. The cluster centers of Kmeans are optimized using random initialization and EM algorithm for at most 300 iterations.\footnote{\url{ https://scikit-learn.org/stable/modules/generated/sklearn.cluster.KMeans.html}} We use RMSprop~\citep{tieleman2012divide} to optimize NNSC for 2,000 iterations.


\textbf{PPLM} uses the default hyperparameters for conditioning on a bag of words in its GitHub repository\footnote{\url{https://github.com/uber-research/PPLM}}. We try several different hyperparameters in PPLM and also try to apply \textbf{PPLM} to the original GPT2 with 117M parameters and to the GPT2 that is fine-tuned on Wikipedia. They produce similar relevancy and perplexity, which are significantly worse than ours in automated evaluation.

The code of \textbf{PPLM} can only condition on a single word piece, so we need to remove the rare words that contain multiple word pieces. We filter out the input prompt in the test set if \textbf{PPLM} cannot condition on any word in the randomly sampled topics.




\subsection{Automated Evaluation}

Similar to training, we first randomly sample a word piece sequence with a length of 512 in the testing set and call the sequence a paragraph. We randomly choose a number from 1 to 79 as the number of word pieces that the first input prompt include and append 80 more word pieces to create the next input prompt until all the word pieces in the paragraph are added to the prompt. When we compute Sim Short in Table \refoptauto, only the first input prompt in the paragraph is used, while all prompts are included when we compute Sim.


In every automatic evaluation, we sample 300 paragraphs. We do not train our model using <|startoftext|> or <|endoftext|> because a paragraph might not start with the beginning of the first sentence, and a paragraph might contain multiple Wikipedia pages.  The maximal input size of our conditional text generator is 256, and it needs to generate 50 word pieces, so we only consider the last 206 word pieces when the input prompt is long. For each input prompt, we sample 3 sentences using our conditional text generator or \textbf{PPLM}.

When computing Sim, Sim Short, Sim Diff, BLEU, and BLEU Diff, we remove the first word piece in the continuation and last word piece in the input prompt because the word pieces might not form complete words in the evaluation. Furthermore, we ignore the input prompt in the test set if the length of continuation in the paragraph is smaller than $O=O'=25$. When computing Dist-1 and Dist-2, we count unigram and bigram within each paragraph.



\subsection{Human Evaluation}
In STSb, we discard the sentences containing less than $K=10$ words after removing stop words to ensure that \textbf{Kmeans-local} could generate 10 non-repetitive topics. 

GPT2 fine-tuned on English Wikipedia sometimes generate sentences containing special characters (e.g., UTF-8 characters for other languages), which crowdsourcing workers might not understand. Thus, we filter out the input prompt in the STSb for human evaluation if the input prompt or the continuation generated by any method contains a character that cannot be encoded using the ASCII code.


On Amazon Mechanical Turk (MTurk), we prepare one task to evaluate the option generators and another task to evaluate the conditional text generators. In the first task, we show the input prompt and the $K=10$ topics generated by a method. Before seeing the generated continuation, the worker needs to answer 
\begin{itemize}[noitemsep,topsep=0pt]
\item "Which topics do NOT promote the narrative?" (TP), and 
\item "Which topics are NOT very likely to appear in the reasonable continuations?" (L).\footnote{We reverse the question because there are often more topics that are likely to appear.} 
\end{itemize}
Then, we show the generated continuation and ask 
\begin{itemize}[noitemsep,topsep=0pt]
\item "How fluent is the generated continuation? (Not fluent at all - Very fluent)" (F), 
\item "How helpful is this generated continuation in terms of promoting the narrative? (Not helpful at all - Very helpful)" (NP), and 
\item "Overall, how good is the generated continuation? (Terrible - Excellent)" (A). \end{itemize}

In the second task, we show the input prompt and the generated continuation. The worker needs to answer 
\begin{itemize}[noitemsep,topsep=0pt]
\item "How fluent is the generated continuation? (Not fluent at all - Very fluent)" (Fluency), and 
\item "Whether the sentence is related to the specified topics?" (Relevancy).
\end{itemize}

We allow only masters on MTurk (the worker with a good reputation) to do our tasks. The workers are rewarded 0.4 or 0.5 dollars for each of the first tasks and 0.2 dollars for each of the second tasks.

In our instruction, we define the reasonable continuation as what the author might say next given only the input prompt, and what the author said in the real word is not important.


The average performance of generated text is between the score 3 and 4. That is, the quality of generated sentences are between somewhat fluent and fluent (F), somewhat helpful and helpful (NP), and medium and good (A). The results suggest the difficulty of generating the continuation for a sentence (mostly from the news in the filtered STSb). 

\section{More Examples}
\label{sec:more_example}

We randomly select 8 examples with less than 130 letters from STSb as our input prompts. The topics of different option generators are visualized in Table~\ref{tb:generated_more_topics}. The continuations of different text generators are visualized in Table \ref{tb:generated_more_text_examples}. You can download our code from \url{https://github.com/iesl/interactive_LM} and test our models using your own prompts via IPython notebook.

\end{document}